\documentclass[12pt]{article}
\usepackage{amssymb,tabularx,multirow,amsmath,natbib, epsfig, threeparttable, amstext}
\usepackage{setspace}
\usepackage{verbatim}
\usepackage{times}
\usepackage{media9} 
\usepackage{algorithm2e}
\usepackage{floatrow}
\usepackage{float}
\usepackage{framed}
\usepackage{caption}
\usepackage{subcaption}
\usepackage{hyperref}

\newcommand{\mbf}[1]{\mathbf{#1}}
\newcommand{\mbv}[1]{\mbox{\boldmath$#1$\unboldmath}}

\def\bh{\mathbf{h}}

\def\bx{\mathbf{x}}

\def\bV{\mathbf{V}}
\def\bU{\mathbf{U}}
\def\bW{\mathbf{W}}

\def\bY{\mathbf{Y}}
\def\bZ{\mathbf{Z}}

\newcommand{\bfbeta}{\mbox{\boldmath $\beta$}}

\begin{document}

\title{\bf  An Ensemble Quadratic Echo State Network for Nonlinear Spatio-Temporal Forecasting}
\author{Patrick L. McDermott\thanks{Department of Statistics, University of Missouri, 146 Middlebush Hall,
Columbia, MO 65211 USA; E-mail: plmyt7@mail.missouri.edu (corresponding author)} \hspace{5mm} Christopher K.
Wikle \\ Department of Statistics \\ University of Missouri}
 \maketitle
 
\begin{abstract}
Spatio-temporal data and processes are prevalent across a wide variety of scientific disciplines. These processes are often characterized by nonlinear time dynamics that include interactions across multiple scales of spatial and temporal variability.  The data sets associated with many of these processes are increasing in size due to advances in automated data measurement, management, and numerical simulator output.   Nonlinear spatio-temporal models have only recently seen interest in statistics, but there are many classes of such models in the engineering and geophysical sciences.  Traditionally, these models are more heuristic than those that have been presented in the statistics literature, but are often intuitive and quite efficient computationally.  We show here that with fairly simple, but important, enhancements, the echo state network (ESN) machine learning approach can be used to generate long-lead forecasts of nonlinear spatio-temporal processes, with reasonable uncertainty quantification, and at only a fraction of the computational expense of a traditional parametric nonlinear spatio-temporal models.  
 \end{abstract}


\newpage

\section{Introduction}\label{sec:Intro}

Spatio-temporal processes in the natural world often exhibit nonlinear behavior such as growth through time, frontal boundaries, density dependence, shock waves, repulsion, nonlinear advection, and predator-prey interactions, to name a few.  Although at some scales in space and time linear models can be quite effective for predicting these processes, it is sometimes essential to consider the nonlinearity explicitly, especially when forecasting at very short or very long time scales, or when interpolating data where there are large gaps and active dynamical processes (e.g., ocean eddies).  Indeed, nonlinearity is often the cause of marginal non-Gaussianity and extremes in such data.   Historically, there has not been much development on parametric models for nonlinear dynamic spatio-temporal processes in the statistics literature, but it has recently seen increasing interest  (e.g., see \cite{wikle2010general}, \cite{wikle2011polynomial}, \cite{gladish2014physically}, \cite{Richardson2017sparsity}).  These models have emphasized state-dependent transition operators and the importance of quadratic interactions, which can be shown to be a fundamental property of many physical and biological processes.  Although these frameworks are quite general and can be effective, they suffer from a curse-of-dimensionality in parameter space, and require careful attention to reduce the effective number of parameters through mechanistically-motivated (hard) shrinkage, regularization, and/or state-reduction (see \cite{wikle2015modern} for an overview).  In addition, even when one controls for the large number of parameters in these models, they can still be quite expensive computationally.   Thus, it is of interest to consider so-called ``black-box'' parametric approaches for spatio-temporal models that can accommodate nonlinearity, while retaining computational efficiency.  

The growth of ``deep learning'' methods in the machine learning literature suggests that these approaches may be suitable as an efficient ``black box'' model to accommodate nonlinear spatio-temporal dynamics.  Although the standard feed-forward neural network is not able to account for the time dependence present in such processes, the recurrent neural network (RNN) approach developed in the engineering literature in the 1980s (e.g., see the review in \cite{lukovsevivcius2009reservoir}), and popularized more recently in machine learning, are designed to allow cycles and sequences in their hidden layers.  These methods have not been used much for spatio-temporal prediction, but they have been used extensively in natural language processing applications, where the sequence of words, and hence their temporal dependence, are fundamentally important.

\begin{figure}[h]
  \centering
   \captionsetup{font=footnotesize}
\includegraphics[height=5cm]{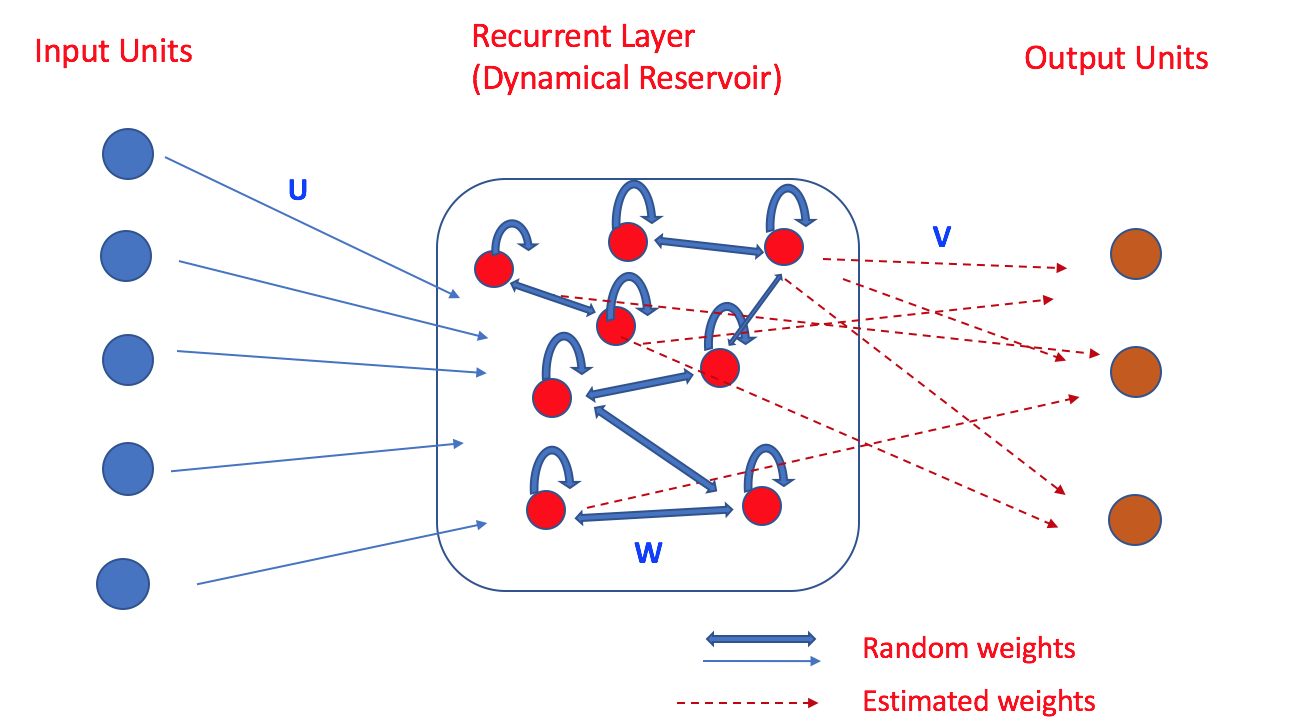}
\caption{Schematic diagram showing the basic echo state machine. The inputs are fed (via weights $\bU$) into a dimension expanded recurrent layer (the dynamical reservoir) with sparse interactions given by the weights $\bW$.  The hidden units from this reservoir are then mapped into the output layer through weights $\bV$.  Importantly, only these output weights (given by the dashed red lines) are estimated.  The input and reservoir weights (given by thin and thick blue lines, respectively), are fixed, sparse, and (importantly) randomly specified. }  
\label{fig:ESNdiagram}
\end{figure}

Traditionally, RNN models have been fairly difficult to fit in settings such as the typical spatio-temproal forecasting problem due to the so-called vanishing gradient problem in the back-propagation algorithm used to obtain weights in the hidden layers.   Because of this, two varieties of RNNs have been developed to minimize the number of weights that need to be trained, so-called echo state networks (ESNs; \cite{jaeger2007echo}) and liquid state machines (LSMs; \cite{maass2002real}).  These approaches, which are now often labeled more generally as  ``reservoir computing'' methods, consider sparsely connected hidden layers that allow for sequential interactions.  In addition, a crucial component of such reservoir models is that the connectivity and the weights for the hidden units are {\it fixed} yet {\it randomly assigned}!  That is, the input data goes into a hidden fixed ``reservoir'' that contains sequential linkages.  This reservoir is typically of higher dimension than the input, so there is a dynamical expansion of the input, thus adding model flexibility.  The reservoir states are then mapped to the desired output, and importantly, {\it only the weights at this mapping phase are estimated}.  Figure \ref{fig:ESNdiagram} shows a schematic of a typical ESN.

ESN models have not been used extensively for spatio-temporal applications, nor have they been considered from a statistical perspective for such processes.
Statistical forecast methods should present reasonable uncertainty metrics for the forecasts they generate. Traditional ESN models, along with most extensions of the ESN model, do not typically provide formal uncertainty measures on the forecasts.  However, there have been a few attempts to quantify uncertainty through the consideration of ensembles or bootstrap samples utilizing different random reservoir weights (e.g., \cite{yao2013ensembles}, \cite{sheng2013prediction}). Despite the introduction of these uncertainty quantification methods, many of the recent developments in the ESN literature have continued to only use point estimate forecasts.  In addition, other heuristic nonlinear forecasting methods (e.g., analog forecasting -- see \cite{mcdermott2016model}, for a recent example) often utilize embeddings, in which lagged values of the inputs are used in the forecast, as motivated by the theorem of \cite{takens1981detecting}.  This is far from standard practice in the ESN literature.  Furthermore, ESN methods do not typically consider quadratic interactions in the mapping between the reservoir and the output.   

We show here that a simple ensemble ESN, with embedded inputs and quadratic reservoir-to-output interactions, can provide very effective forecasts (with uncertainty measures) for notoriously difficult nonlinear spatio-temporal dynamical systems.  In particular, we consider a simulation experiment with the classic \cite{Lore:96} 40-variable nonlinear system, and a real-world example for long-lead forecasting of tropical Pacific sea surface temperature (SST), an important problem due to the importance of the El Ni\~no and La Ni\~na phenomena that form in the region, thus significantly impacting weather conditions across the globe. 

Section 2 describes our ensemble quadratic spatio-temporal ESN model, followed by simulation results in Section 3 based on the \cite{Lore:96} nonlinear ordinary differential equation system.  We then present a real-world application of the approach for long-lead (six month) forecasting of tropical Pacific SSTs in Section 4.  Section 5 offers a brief discussion and conclusion, including suggestions for future research.

\section{Spatio-Temporal Echo State Network (ESN)}\label{sec:STESN}

A very basic RNN (e.g., \cite{lukovsevivcius2009reservoir}) can be specified as follows for time $t=1,\ldots,T$:
\begin{eqnarray}
\mbox{response: }  & \; & {\mbf Y}_t = g_o({\mbf o}_t) \nonumber \\
\mbox{output: } & \; & {\mbf o}_t = {\mbf V} {\mbf h}_t \label{eq:V}\\
\mbox{hidden state: } & \; & \bh_t = (1 - \alpha)\; \bh_{t-1} + \alpha \; \widetilde{\bh}_t, \label{eq:h}\\
& \; & \widetilde{\mbf h}_t = g_h({\mbf W} {\mbf h}_{t-1} + {\mbf U}{\mbf x}_t), \label{eq:htil} 
\end{eqnarray}
where ${\mbf Y}_t$ is an $n_y$-vector of responses at time $t$, ${\mbf x}_t$ is an $n_x$-dimensional input vector (typically, assumed to include a one in the first position for an ``intercept''), ${\mbf o}_t$ is an $n_y$-vector of outputs that are associated with a linear transformation of the $n_h$-dimensional hidden unit vector, ${\mbf h}_t$ (with $\widetilde{\bh}_t$ its update), ${\mbf W}$, ${\mbf U}$ are $n_h \times n_h$ and $n_h \times n_x$ hidden layer weight matrices, respectively, ${\mbf V}$ is the $n_y \times n_h$ output weight matrix, and $g_o(\cdot)$, $g_h(\cdot)$ are specified activation functions.  The $\alpha$ parameter in (\ref{eq:h}) takes a value $(0,1]$ and is known as the ``leaking rate.''  The ESN version of this simple RNN then considers the hidden layer weight matrices ${\mbf W}$ and ${\mbf U}$ (the reservoir weights) to be fixed; they are just drawn once from a distribution centered around zero, with added sparsity (see Section \ref{sec:QESN} below for details).  Only the output matrix ${\mbf V}$ is estimated!  This presents a huge computational cost savings since there are relatively few output weight parameters, and they can be estimated through standard regularization-based statistical estimation approaches (e.g., if $g_o(\dot)$ is the identity function, then a simple ridge regression estimation of ${\mbf V}$ is typically used). The ESN model has gained much popularity in large part due to this computational advantage.

It is important to understand the role of the reservoir given by equation (\ref{eq:htil}) in the basic ESN framework.  As described in \cite{lukovsevivcius2012practical}, the hidden units in the reservoir act as a nonlinear expansion of the input vector, $\bx_t$, and perhaps more importantly, as a way to establish ``memory'' or account for the sequential nature of the dependence in the input vectors and, ultimately, the response.  The idea of a nonlinear expansion in a high dimension helps to magnify potentially salient dynamic features of the input, and the output weights ($\bV$ in equation (\ref{eq:V})) provide a way to select those expanded states that are important for the response. 

In Section \ref{sec:QESN} we describe this model in more detail and include our modifications for spatio-temporal prediction.

\subsection{Quadratic ESN}\label{sec:QESN}

As described in Section \ref{sec:Intro}, when predicting nonlinear spatio-temporal processes it is often quite important to include quadratic interactions between hidden processes and the response, as well as embeddings (lagged values) of the input.  Very simple modifications to the basic ESN of Section \ref{sec:STESN} allow for these important model components.  Below we show one such modified model, a basic quadratic ESN (QESN) for continuous output (i.e., where $g_o(\cdot)$ is the identity function).  We note that it is straightforward to include nonlinear activation functions at this stage (e.g., softmax) depending on the response support and the goal of the analysis (i.e., classification versus regression).

\subsubsection{Spatio-Temporal Quadratic ESN}

\vspace{2mm}

For $t=1,\ldots,T$, the QESN model is given by:
\begin{eqnarray}
\mbox{response: }  & \; & {\mbf Y}_t = {\mbf V}_1 {\mbf h}_t + {\mbf V}_2 {\mbf h}^2_t + {\mbv \epsilon}_t, \;\; {\mbv \epsilon}_t \; \sim \; N({\mbf 0},{\mbf R}), \label{eq:QESNresp}\\
\mbox{hidden state: } & \; & \bh_t = (1 - \alpha) \; \bh_{t-1} + \alpha \; \widetilde{\bh}_t, \label{eq:QESNhidden2}\\
 & \; &  \widetilde{\mbf h}_t = g_h\left(\frac{\nu}{|\lambda_w|}{\mbf W} {\mbf h}_{t-1} + {\mbf U}\tilde{\mbf x}_t\right), \label{eq:QESNhidden} \\
\mbox{parameters: } & \; & {\mbf W} = [w_{i,\ell}]_{i,\ell}: w_{i,\ell} = \gamma^w_{i,\ell} \; \text{Unif}(-a_w,a_w) + (1 - \gamma^w_{i,\ell}) \; \delta_0, \label{eq:QESNw} \\
& \; & {\mbf U} = [u_{i,j}]_{i,j}: u_{i,j} = \gamma^u_{i,j} \; \text{Unif}(-a_u,a_u) + (1 - \gamma^u_{i,j}) \; \delta_0, \label{eq:QESNu}  \\
& \; & \gamma_{i,\ell}^w \; \sim \; Bern(\pi_w) \nonumber \\
& \; & \gamma_{i,j}^u \; \sim \; Bern(\pi_u) \nonumber,
\end{eqnarray}
where ${\mbf Y}_t$ is the $n_y$-dimensional response vector at time $t$, ${\mbf h}_t$ is the $n_h$-dimensional hidden state vector, and 
\begin{equation}
\tilde{\mbf x}_t = [\bx'_t,\bx'_{t-\tau}, \bx'_{t- 2 \tau},\ldots,\bx'_{t - m \tau}]',
\label{eq:QEQNembed}
\end{equation}
is the $n_{\tilde{x}} = (m+1) n_x$-dimensional embedding input vector, containing embeddings for time periods $t- \tau$ through $t - m \tau$, where $\tau$ is often the forecast lead time (although, generally $\tau$ can be any integer).  Furthermore, $\delta_0$ is a Dirac function at zero, and $\lambda_w$ corresponds to the largest eigenvalue of ${\mbf W}$ (i.e., the ``spectral radius'' of $\bW$).  The only parameters that are estimated in this model are ${\mbv V}_1$ and ${\mbf V}_2$, and ${\mbf R}$ from equation (\ref{eq:QESNresp}), for which we use a ridge penalty hyper-parameter, $r_v$.   Importantly, note that the ${\mbf W}$ and ${\mbf U}$ matrices are simulated from a mixture distribution of small values (uniformly sampled in a range from between $(-a_w,a_w)$ and $(-a_u,a_u)$) with, on average, $(1-\pi_w)$ and $(1-\pi_u)$ elements set equal to zero, and then assumed to be fixed.  The hyper-parameters $\{\nu, n_h, r_v, \pi_w, \pi_u, a_w, a_u\}$ are specified (see below).

\subsection{Ensemble QESN}\label{sec:EnsQESN}
Most traditional ESN applications do not include a mechanism to quantify the uncertainty of the model predictions.  This is perhaps somewhat surprising given that the reservoir weight parameters are not estimated, but are chosen at random.  We would expect that the model is likely to behave differently with a different set of $\bW$ and $\bU$ weights.  This is especially true when the number of hidden units is fairly small.   Although traditional ESN models have a large number of hidden units, it can be desirable to have many ensemble members with smaller number of units.   This provides flexibility in that it prevents overfitting, allows the ensemble members to behave as a committee of relative weak learners, and gives a more realistic sense of the prediction uncertainty for out-of-sample forecasts. 
Thus,  we could generate an ensemble or bootstrap sample of forecasts (e.g., \cite{yao2013ensembles}, \cite{sheng2013prediction}).  This ensemble approach can easily be implemented with our model as given in Algorithm 1 (note, R code is included in the online supplemental materials), to make out-of-sample forecasts for $n_f$ periods.

\vspace{3mm}

\begin{algorithm}[H]
 \KwData{$\{\bY_t$, $\tilde{\bx}_t$: $t=1,\ldots,T\}$}
 \KwResult{Ensemble of predictions: $\{\widehat{\bY}^k_t: t=T+1,\ldots,T+n_f; k=1,\ldots,K\}$ }
 
 Initialize: Select tuning parameters $\{n_h, \nu, r_v, \pi_w, \pi_u, a_w, a_u, \alpha \}$ (e.g., by validation or cross-validation)   \;
 \For{$k=1,\ldots,K$}{
 Simulate $\bW^k$, $\bU^k$ using (\ref{eq:QESNw}) and (\ref{eq:QESNu})
 
 Calculate $\{\bh_t^k: t=1:T\}$ using (\ref{eq:QESNhidden2}) and (\ref{eq:QESNhidden})
 
 Use ridge regression to estimate $\bV^k_1$, $\bV^k_2$
 
 Calculate out-of-sample forecasts $\{\widehat{\bY}^k_t: t=T+1,\ldots,T+n_f\}$ (requires calculating $\{\widehat{\bh}^k_t: t=T+1,\ldots,T+n_f\}$ from the reservoir) 
  }
  
  Use ensemble of forecasts $\{\widehat{\bY}^k_t: t=T+1,\ldots,T+n_f; k=1,\ldots,K\}$ to calculate moments, prediction intervals, etc.

\vspace{3mm}

 \caption{Simple Ensemble-QESN Algorithm}
\end{algorithm}

\subsection{Model Parameterizations and Hyper-Parameters}

The model presented in Section \ref{sec:QESN} depends on several hyper-parameters, some of which are typically more important than others in ESN applications.  An excellent and detailed summary of the practical issues associated with traditional ESN implementation is given in \cite{lukovsevivcius2012practical}.  We discuss some of these notions here, along with our experience in the context of spatio-temporal forecasting with this specific model. 

The size of the reservoir, $n_h$, is traditionally one of the more important hyper-parameters.  In most implementations, one seeks a reservoir with a large number of hidden states ($n_h$ large) and assumes that regularization will mitigate the potential to overfit.   A rule of thumb in traditional ESN settings is to make $n_h$ as large as possible, so that $T > n_x + n_h$.  In the spatio-temporal context, we typically do not have extremely large values of $T$, but we make up for that with the embedding input, $\tilde{\bx}_t$.   Thus, we have found in our settings that smaller values of $n_h$ are often sufficient.  Also, as described in Section \ref{sec:EnsQESN}, we generally prefer $n_h$ to be relatively small so that our ensemble acts more as a committee of weak learners, which helps prevent against overfitting.  In practice, we typically select $n_h$ through a validation or cross-validation procedure (see Section \ref{sec:Lorenz} and \ref{sec:SST}).  

In traditional ESN applications, the leaking rate parameter, $\alpha$, is often quite important. We have found that in our QESN model this is not typically the case.  That is, in almost every application we have considered, validation and cross-validation have suggested that $\alpha = 1$ is the best setting for this parameter, in which case, $\widetilde{\bh_t} = \bh_t$. However, this need not be the case since small leak rates can be helpful for slowly varying systems ( \cite{lukovsevivcius2012practical}). Leakage rates are therefore application dependent and so we recommend evaluating whether $\alpha \neq 1$ improves out-of-sample prediction.

The scaling of the hidden state reservoir weighting matrix, $\bW$, in (\ref{eq:QESNhidden}) is quite important.  In general, the spectral radius (largest eigenvalue) of $\bW$ must be less than one to ensure what is known in the literature as the ``echo state property.''  This is a property that allows, with large enough time increments, for the hidden states to lose their dependence on the initial input conditions.  Practically, when the spectral radius is not less than one, the hidden state can experience complex nonlinear dynamics (e.g., multiple fixed points, periodicities, and chaotic behavior), which destroys the echo state property (e.g., see the discussion in \cite{lukovsevivcius2012practical}).  A rule of thumb is that a smaller spectral radius should be used if the responses are more dependent on the input at recent times and a larger value (but still less than one) should be used if the responses depend more on the past.  For our purposes, this means that the parameter $\nu$ should be quite important as it controls the spectral radius.  That is, dividing $\bW$ by $|\lambda_w|$ gives a spectral radius of one, and so multiplying by $\nu < 1$ gives us the flexibility to control the overall spectral radius of the hidden state weighting parameters.   
We typically choose the specific value of $\nu$ by validation or cross-validation.
 
There is some debate in the literature about the importance of sparsity on the reservoir weight matrices, $\bW$ and $\bU$.  We have found in our applications that it is important that these be quite sparse (e.g., 80-95\% zeros). Both $\pi_w$ and $\pi_u$ can also be selected through validation or cross-validation, although the model is not sensitive to the specific (small) value of either parameter.  While we use the truncated uniform distribution to sample the parameter estimates for these in (\ref{eq:QESNw}) and (\ref{eq:QESNu}) above, we have not found the results to be very sensitive to the choice of this distribution (e.g., uniform versus a standard normal distribution), which is consistent with the traditional ESN literature.

For the applications presented here, the model was moderately sensitive to the choice of ridge parameter, $r_v$, used for estimating $\bV_1$ and $\bV_2$.  We typically use validation or cross-validation to select this parameter.  We note that other forms of regularization (e.g., $L_1$-penalties, or hybrid $L_1, L_2$ penalties) could be used here as well.  

Lastly, it is typically the case that inputs are normalized in ESN applications, although in principle, this can be accommodated through the parameter scaling (i.e., the $a_w$, and $a_u$ hyper-parameters).   Note also that it is known that the reservoir acts to compress the variability of the principal components of the inputs, $\bx_t$.  Thus, it is often recommended that if one uses principal components as inputs, then it may be best to not include those that are associated with the smallest variability, as their importance and influence will be exaggerated in the reservoir  (e.g., \cite{lukovsevivcius2012practical}).  In spatio-temporal applications, we typically use some form of dimension reduction on the input $\bx_t$ vectors (e.g., empirical orthogonal functions (EOFs) -- which are just principal components for spatio-temporal data,  e.g., see \cite{CandW2011}; but, other dimension reduction approaches can be used).   We also typically work with response vectors that have been projected onto their leading EOFs given the large number of spatial locations that are often present in spatio-temporal prediction problems. 

In summary, we have found in our spatio-temporal applications that the results tend to be more sensitive to the choice of $\{n_h, \nu, r_v\}$ than $\{\pi_w, \pi_u, a_w, a_u\}$, but recommend evaluating the sensitivity of the forecasts to these parameters in new applications.

\section{Simulated Data: The 40-Variable Lorenz System}\label{sec:Lorenz}

In order to evaluate the model and the importance of the modifications considered in Section \ref{sec:QESN} (i.e., embedding inputs, quadratic outputs, and ensemble uncertainty quantification), we consider simulated data from the classic 40-variable nonlinear model of \cite{Lore:96}, referred to as the ``Lorenz-96'' model. Although the classic 3-variable Lorenz model is often cited in the ESN literature, to our knowledge, this more spatially relevant 40-variable model has not been evaluated in the ESN context.   In particular, the system is governed by the set of 40 ordinary differential equations given by
\begin{equation}
\frac{dz_{t,i}}{dt} = (z_{t,i+1} - z_{t,i-2})z_{t,i-1} - z_{t,i} + F_t, \;\;\; i=1,\ldots,40,   \label{eq:L96}
\end{equation}
where $F_t$ is a forcing variable, and the variables $z_{i,t}$ correspond to state variables at 40 equally spaced locations on a circular (e.g., latitude circle) one-dimensional spatial domain with periodic boundary conditions (e.g., $z_{41,t} = z_{1,t}$).  The nonlinearity in (\ref{eq:L96}) comes from the quadratic interactions across locations, and this was originally designed to mimic nonlinear advection processes in geophysical fluids.  

The data were simulated by using a simple Euler solver with a time step of $\Delta = 0.10$ and forcing value of $F_t=5$.   Moreover, the lead time is set to 6 periods in order to increase the nonlinearity and create a more realistic forecasting scenario. Gaussian white noise error was added to each realization, so that $Y_{i,t} = z_{i,t} + \eta_{i,t}$, where $\eta_{i,t} \; \sim N(0,\sigma^2_\eta)$ for $i=1,\ldots,40$, where $\sigma_\eta = 0.5$.  We considered 750 time periods (post burn-in), with the last 99 time periods held out for out-of-sample prediction.  

\subsection{Lorenz-96 Model Validation}\label{sec:L96valid}

A validation approach was used to select the hyper-parameters for the QESN model.  Specifically, we consider the following grid of hyper-parameters: $n_h=\{30+15\times q_h : q_h=0,\dots,5 \}$, $\nu=\{.05\times q_\nu : q_\nu=1,\dots,20 \}$, and $r_v=\{ .001,.005,.01\}$. Furthermore, we had little {\it a priori} information about how the input should be embedded, hence we also included the number of embeddings ($m$) as a hyper-parameter in the grid search.  Specifically, using a value of $\tau=1$, which produced more accurate forecasts than using the lead time of 6, the parameter $m$ was evaluated over a grid of $\{2\times q_m: q_m=1,\dots, 5 \}$. We should note that when embeddings have been used in the ESN literature, the embedding dimension is often set heuristically without considering different possible values.  Finally, the model was not very sensitive to the hyper-parameters in the set $\{\pi_w, \pi_u, a_w, a_u\}$ and so all four were set to $0.1$.  To perform the validation we trained the model on the first 551 observations and held out the next 99 observations for the validation set.  All models were evaluated via the mean square error (MSE) calculated over all validation periods and locations, using an ensemble size of 500 (i.e., $K=$500 in Algorithm 1).  The  parameters associated with the lowest validation sample MSE were $n_h=60,\nu=.55,r_v=.001$, and $m=4$.

\subsection{Lorenz-96 Model Out-of-Sample Forecasts}

\begin{figure}[H]
  \centering
  \captionsetup{font=footnotesize}
\includegraphics[width=13cm,height=7cm]{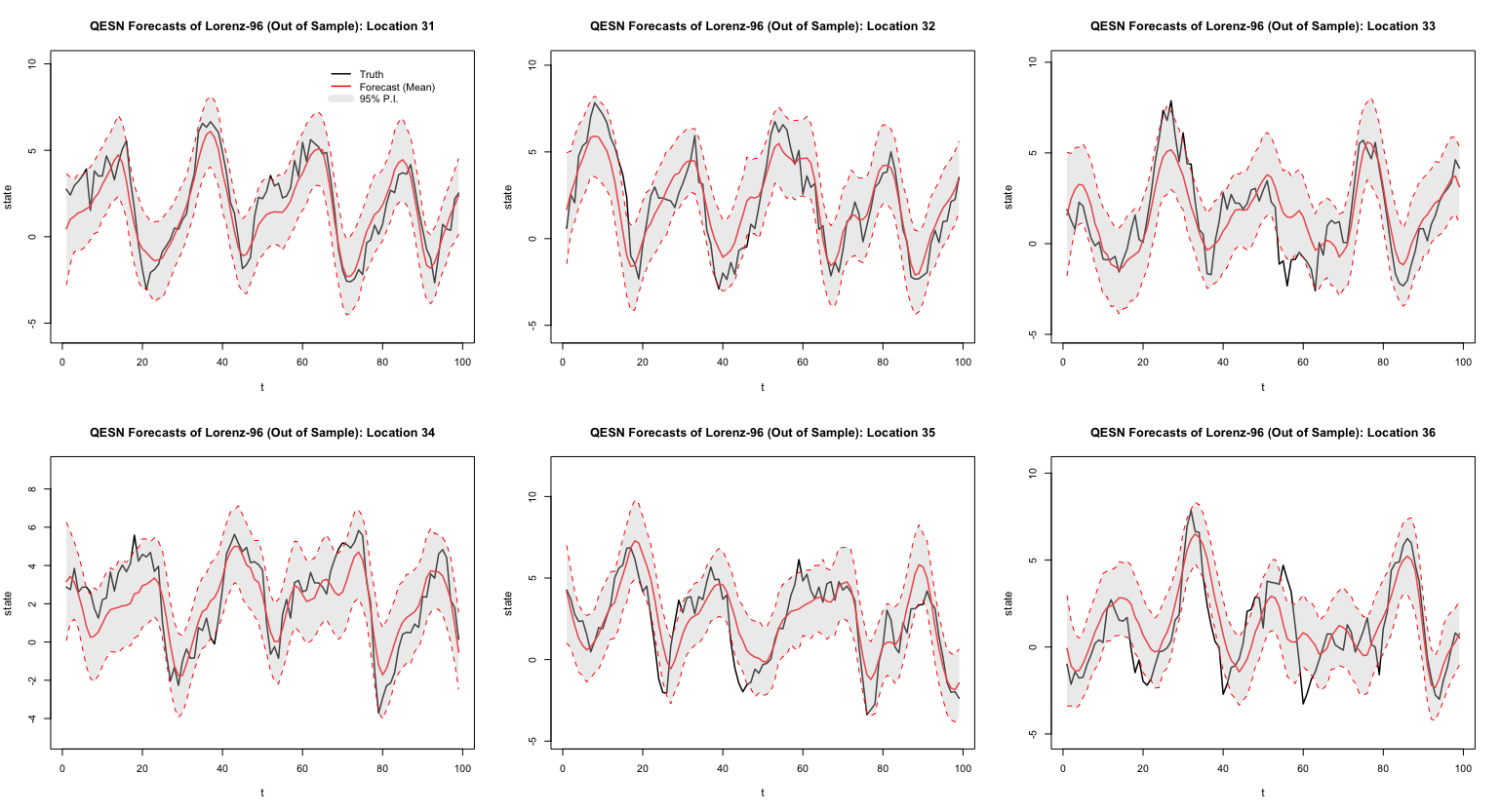}
\caption{Out-of-sample forecasts for six locations from the simulated Lorenz-96 system over 99 periods with a lead time of 6 periods. The black line represents the truth, while the solid red line denotes the forecasted mean from the ensemble QESN model using hyper-parameters found through validation. The shaded grey area signifies the 95\% prediction intervals.}  
\label{fig:L96for}
\end{figure}

Using the validation parameters given in Section \ref{sec:L96valid}, the model was trained using the first 651 observations, with the last 99 observations held out for out-of-sample forecasts.  These forecasts and associated 95\% prediction intervals (P.I.'s) are given in Figure \ref{fig:L96for} for 6 of the 40 locations. Considering the difficulty of forecasting a nonlinear system 6 time periods forward, Figure \ref{fig:L96for} shows that the forecasts for the QESN model generally correctly forecast the pattern of time evolution of the system.   Perhaps more importantly, across all locations the model has strong prediction interval coverage probabilities with 95.4\% of the true values falling within the 95\% P.I.'s. This example demonstrates the potential for the ensemble QESN model to successfully forecast spatio-temporal systems at nonlinear time scales, and to capture reasonable forecast uncertainty.

\section{Application: Long-Lead Forecasting of Pacific SST}\label{sec:SST}

The anomalous warming (El Ni\~no) and cooling (La Ni\~na) of the tropical Pacific ocean that occurs quasi-periodically on time scales of 3-5 years accounts for one of the largest sources of variability in weather systems across the globe.  These phenomena are sometimes collectively known as the El Ni\~no Southern Oscillation (ENSO).  The effects of ENSO variability can be quite serious in terms of heat waves, drought, flooding, and increased potential for other types of severe weather.   For this reason, there has long been interest in forecasting the state of the tropical Pacific ocean surface temperature many months into the future in order to facilitate resource planning.  ENSO is a very complicated multivariate process that operates on many spatio-temporal scales of variability and it is known to exhibit nonlinear evolution at certain time scales.  In addition, many forecast methods that have performed reasonably well in past cycles of ENSO did not work very effectively in the year leading up to the last major ENSO cycle in 2015-2016 (\cite{l2016observing}, \cite{hu2017extreme}).  Thus, we consider the 6-month long lead forecasts of Pacific SST over this period to illustrate our ensemble QESN model.

Long lead prediction is one of the few scenarios in modern climatological and weather forecasting where statistical methods can do as well or better than deterministic methods (i.e., numerical solutions to partial differential equations that govern the ocean and atmosphere), as summarized in \cite{barnston1999predictive},  \cite{jan2005did}.   Although linear models have been shown to produce skillful forecasts of ENSO (e.g., \cite{penland1993prediction}; \cite{knaff1997nino}), it has been established that models that can accommodate nonlinear interactions often work better, particularly in forecasting the evolution of the El Ni\~no phase of the ENSO cycle.   In particular, successful nonlinear models that have been used in the past include the nonlinear analog approach of \cite{drosdowsky1994analog}; classical neural network models of \cite{tangang1998forecasting} and \cite{tang2000skill};  switching Markov model of \cite{berliner2000long}; empirical nonlinear inverse models of \cite{timmermann2001empirical}; empirical model reduction methods of \cite{kondrashov2005hierarchy}, \cite{kravtsov2005multilevel}, \cite{kravtsov2009empirical};  and the general quadratic nonlinear (GQN) models of \cite{wikle2010general}, \cite{wikle2011polynomial}, and \cite{gladish2014physically}.  However, most of these methods were developed outside of statsistics and are heuristic with little or no formal uncertainty quantification.  Exceptions include the switching Markov and GQN models, which were developed in the statistics literature and include a formal (hierarchical Bayesian) uncertainty quantification.  Indeed, one of the strengths of the formal GQN approach is that even when its forecast means are not significantly better than these other approaches, its quantification of uncertainty in terms of prediction intervals tends to be much more realistic for the unusually strong ENSO events.  Yet, these formal hierarchical nonlinear statistical SST forecast models can be quite expensive to run, thus suggesting the possible utility of more efficient models such as the ensemble ESN model presented here. 

\subsection{Data}
The long-lead SST forecasting application consists of data from the publicly available Extended Reconstruction Sea Surface Temperature (ERSST) data provided by the National Ocean and Atmospheric Administration (NOAA) (\url{http://iridl.ldeo.columbia.edu/SOURCES/.NOAA/}). In particular we use monthly data of SST anomalies over a period from 1970-2016. The spatial domain of the data is over $29^{\circ}$S-$29^{\circ}$N latitude and $124^{\circ}$E-$70^{\circ}$W longitude, with a resolution of $2^{\circ} \times 2^{\circ} $ (i.e., 2,229 oceanic spatial locations).  Anomalies were calculated by subtracting the monthly climatological means calculated over the period 1981--2010.

\begin{figure}[H]
  \centering
    \captionsetup{font=footnotesize}
\includegraphics[width=9cm]{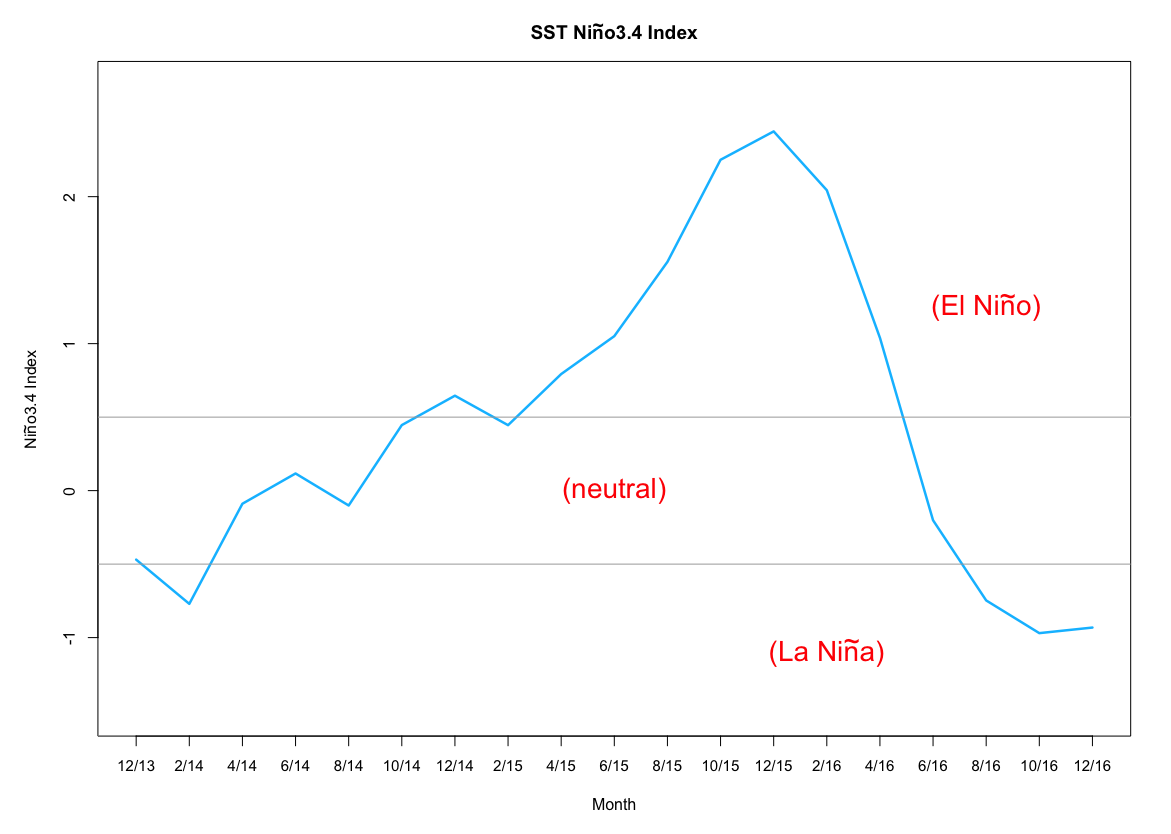}
\caption{Time series plot of the average SST for all grid locations in the Ni\~no 3.4 region ($5^{\circ}$S-$5^{\circ}$N,$120^{\circ}$-$70^{\circ}$W)  from December 2013 through December 2016. For purposes of discussion, values above $0.5$ represent El Ni\~no periods, and periods falling below $-0.5$ are considered La Ni\~na periods. }
\label{fig:N34plot}
\end{figure}

\begin{figure}[H]
  \centering
      \captionsetup{font=footnotesize}
\includegraphics[width=8cm,height=12cm]{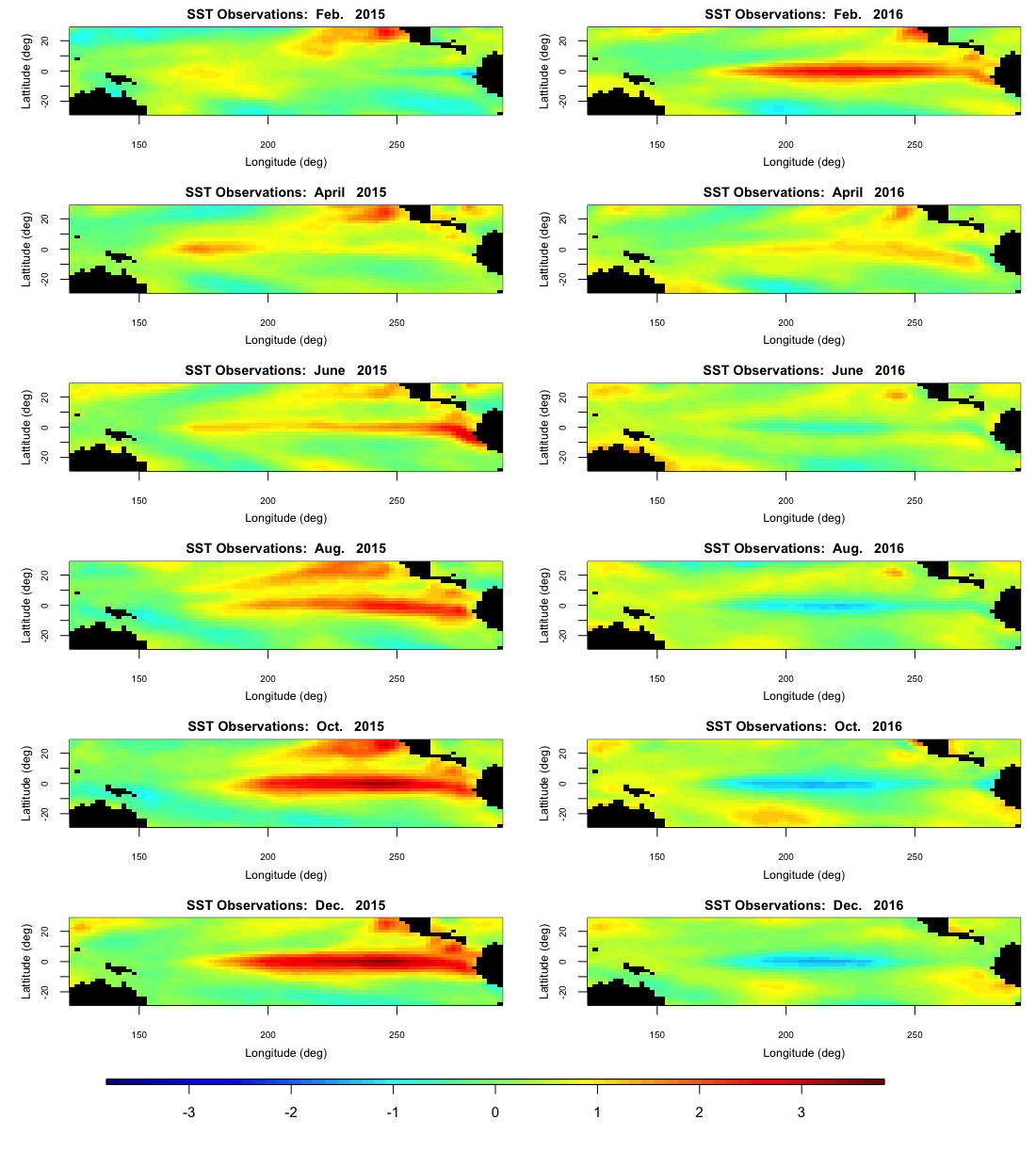}
\caption{Spatial maps of SST anomalies for February 2015 through December 2016. The temporal evolution of these maps represent the start of an El Ni\~no in 2015 and a fairly rapid transition into a La Ni\~na during the latter part of 2016.}  
\label{fig:SSTplots}
\end{figure}

The evolution of SSTs associated with the 2015--2016 ENSO is illustrated in Figure \ref{fig:SSTplots}, which shows gridded spatial fields of SST anomalies from February 2015 through December 2016.  These panels show a relatively fast transition from an El Ni\~no event (warmer than normal anomalies) into a La Ni\~na event (colder than normal anomalies), where the evolution of the two events differs greatly. This difference in the evolution of the onset of these two events is indicative of the different degrees of nonlinearity that are present in the ENSO phenomena, thus presenting a valuable sequence of events for evaluating the QESN model. That is, it is often the case that the development of the warm phase is more nonlinear than the development of the cold phase.  The ENSO trajectory can be seen more concisely by using the common univariate summary measure for ENSO, the Ni\~no 3.4 index, which represents the average of the SST anomalies over the so-called Ni\~no 3.4 region ($5^{\circ}$S-$5^{\circ}$N,$120^{\circ}$-$70^{\circ}$W).  The Ni\~no 3.4 index time series for the period from December 2013 - December 2016 is plotted in Figure \ref{fig:N34plot}. Due to the overall importance of the Ni\~no 3.4 region, it is common to use this index to evaluate long-lead SST forecasts (\cite{barnston2012skill}).

\subsection{Ensemble QESN Implementation} 

To select the hyper-parameters for the QESN model we conducted a validation study by using periods from the ENSO event that occurred from 1997--1999 as a holdout sample. In particular, all of the data from January 1970 though December 1996 were used to train the QESN model, while a sequence of months from May 1997 through August 1999 made up the holdout sample. Both the output and the embedded input consisted of principal component time series associated with the first 10 EOFs (note, spatial fields correspond to the product of these time series and the EOF basis functions; these spatial fields are used for plotting and for calculating the Nin\~no 3.4 index average of the forecast). If $\bZ_t$ denotes the entire spatial field (output) at period $t$, then using EOFs the field can be written as follows: $\bZ_t  	\approx {\mbf \Psi} \bfbeta_t$, where  ${\mbf \Psi}$ is a matrix of EOF basis functions and $\bfbeta_t$ is a vector of expansion coefficients. Therefore, the expansion coefficients can be used as the output (i.e., $\bY_t$) in the QESN model. A similar decomposition can be utilized for the input.

The validation study uses the same grid of values for the $n_h, \nu,$ and $r_v$ parameters as the Lorenz example.  In addition, $\tau$ is set to the lead time (i.e., 6 months) for the embedded input vectors and $m$ was varied over the set $\{1,2,3,4,5\}$. Once again, we found that the model forecasts were not very sensitive to the particular value for the hyper-parameters $\{\pi_w, \pi_u, a_w, a_u\}$ and so we used the same values as in the Lorenz example. Evaluation of a particular set of hyper-parameters was based on the MSE of the Ni\~no 3.4 region average, calculated by using Algorithm 1 with 500 ensembles.  The lowest MSE for the validation study was associated with the following hyper-parameters: $n_h=120,\nu=.35, r_v=.01$, and $m=4$.  

\begin{figure}[H]
  \centering
    \captionsetup{font=footnotesize}
\includegraphics[width=10cm,height=10cm]{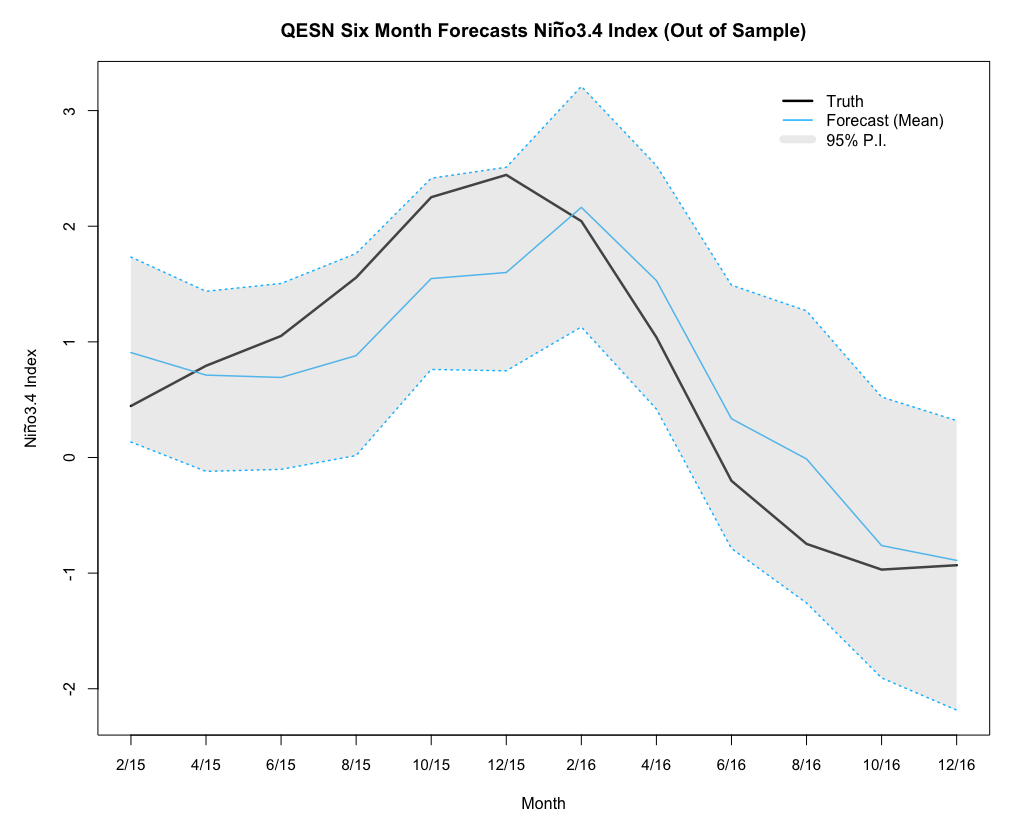}
\caption{Out-of-sample 6-month lead time forecasts and prediction intervals for every two months from February 2015 -- December 2016 for the average SST anomalies in the Ni\~no 3.4 region. The solid black line represents the truth calculated from the data and the corresponding forecast mean from the ensemble QESN is denoted by the solid blue line. The shaded grey area represents 95\% prediction intervals from the ensemble QESN model.}  
\label{fig:Figure_5}
\end{figure}

Evaluation of the ensemble QESN model was conducted by making out-of-sample forecasts for the 2015--2016 ENSO cycle. That is, after training the model from January 1970 --  September 2015 using the hyper-parameters found in the validation study, 6 month lead time out-of-sample forecasts were made for every two months of the 2015--2016 ENSO cycle from February 2015 -- December 2016 (as show in Figure \ref{fig:Figure_5}).  After validation,  the R-program version of this model (see the online supplement) took less than 15 seconds to train and generate predictions using a 2.3 GHz laptop.

\subsection{ Ensemble QESN Results}
As previously discussed, forecasting the Ni\~no 3.4 region is both vital from a planning and management perspective as well a challenging forecasting problem. Forecast means of the Ni\~no 3.4 index during the 2015-2016 ENSO cycle, along with their 95\% P.I.'s, are shown in Figure \ref{fig:Figure_5}. Overall, the QESN model produces uncertainty bounds that cover the observed Ni\~no 3.4 index for the entire 2015-2016 ENSO cycle. Similar to almost every other statistical and deterministic model, the QESN model underestimated the peak of the cycle during December 2015, yet the uncertainty bounds appear very realistic (and, superior to many of those used for operational forecasts). 

\begin{figure}[H]
  \centering
    \captionsetup{font=footnotesize}
\includegraphics[width=10cm,height=10cm]{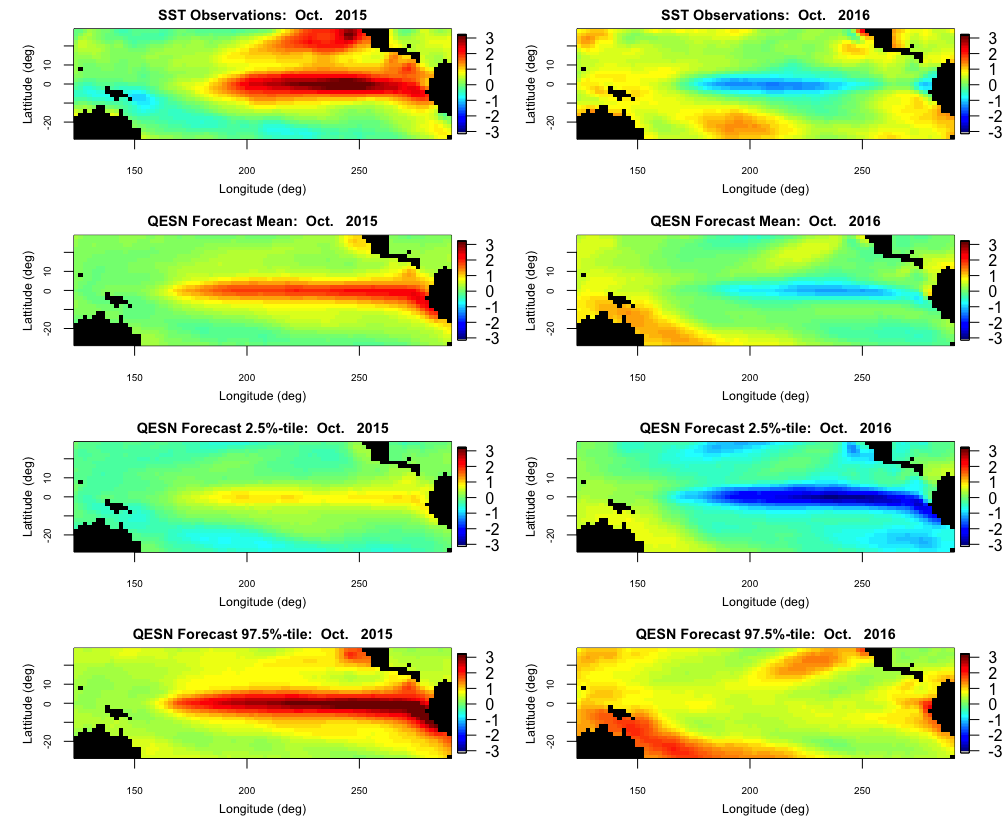}
\caption{Forecast summary maps for all spatial locations for 6-month forecasts valid on October 2015 and  October 2016. The top row denotes the true SST anomalies for the respective years. Forecasted means from the ensemble QESN model for each year are given in the second to top row. The bottom two rows capture the lower and upper quantiles for a 95\% prediction interval calculated in each grid box for each year.}  
\label{fig:Figure_6}
\end{figure}

Although the Ni\~no 3.4 index can provide valuable information and is a useful summary of forecast accuarcy, there is also an interest in examining model forecast performance over the entire spatial domain. Out-of-sample 6-month forecasts for all 2,229 grid locations are shown in Figure \ref{fig:Figure_6} for October 2015 and 2016. For both time periods, the model forecast largely picks up the intensity of the phenomenon, while also capturing much of the true values within the 95\% prediction intervals.  This later point is very important, as many of the heuristic forecast methods in use do not provide a formal uncertainty quantification.

To examine the usefulness of the embedding and quadratic extensions of the ESN model, three additional models were run with the hypar-parameters selected in the validation study. In particular, we ran a model: without any embeddings (labeled M1), without any quadratic terms (labeled M2), and without both embeddings and quadratic terms (labeled M3). Along with evaluating these models with the overall MSE (over all locations) and the Ni\~no 3.4 index MSE, we also considered the continuous ranked probability score (CRPS) over all locations. The CRPS summary is useful for evaluating the prediction accuracy of a forecast, while also considering the distribution of the forecast, and thus the quantification of uncertainty (\cite{gneiting2014probabilistic}). Across all three summary metrics displayed in Table \ref{table_1}, the ensemble QESN model clearly out-performs the other models, suggesting that embedding inputs and quadratic output components are helpful in producing better forecast distributions.  

\begin{table}[H]
\begin{framed}
\captionsetup{font=footnotesize}
\begin{tabular}{ccccccccc} 
Model & Overall MSE  & Ni\~no 3.4 MSE &  CRPS  \\
\hline
QESN&  {\bf 0.288}  & {\bf .261} & {\bf 3.722} \\ 
M1 &  0.343 & 0.545 & 4.570   \\ 
M2 & 0.328 & 0.586 & 4.595   \\ 
M3 & 0.345 &  0.741 & 4.845  \\ 
\end{tabular}
\caption{Results for the ensemble QESN and models M1--M3 in terms of mean squared error (MSE) over all 2,229 locations (Overall MSE), the MSE for the Ni\~no 3.4 region (Ni\~no 3.4 MSE), and the continuous ranked probability score (CRPS) over all locations.  M1 denotes a model without any embeddings, M2 is a model without any quadratic output terms, and M3 is a model without embeddings or quadratic output terms. }
\label{table_1}
\end{framed}
\end{table}

Finally, it is useful to compare the ensemble QESN model to the general quadratic nonlinear (GQN) model of \cite{wikle2010general}, which has been shown to be a useful nonlinear dynamical spatio-temporal model that incorporates formal uncertainty quantification.  In particular, we compared the 6-month lead time forecast distributions from the ensemble QESN model and the GQN model for forecasts valid in October 2015 and October 2016 (near the most intense portions of the El Ni\~no and La Ni\~na in this ENSO cycle).  Note that while it took less than 15 seconds to generate forecasts with the ensemble QESN model, it took over 1 hour to generate an equivalent number of posterior samples from the GQN model!  

\begin{figure}[H]
  \centering
    \captionsetup{font=footnotesize}
    \begin{tabular}{cc}
   \includegraphics[width=4.5cm,angle=-90]{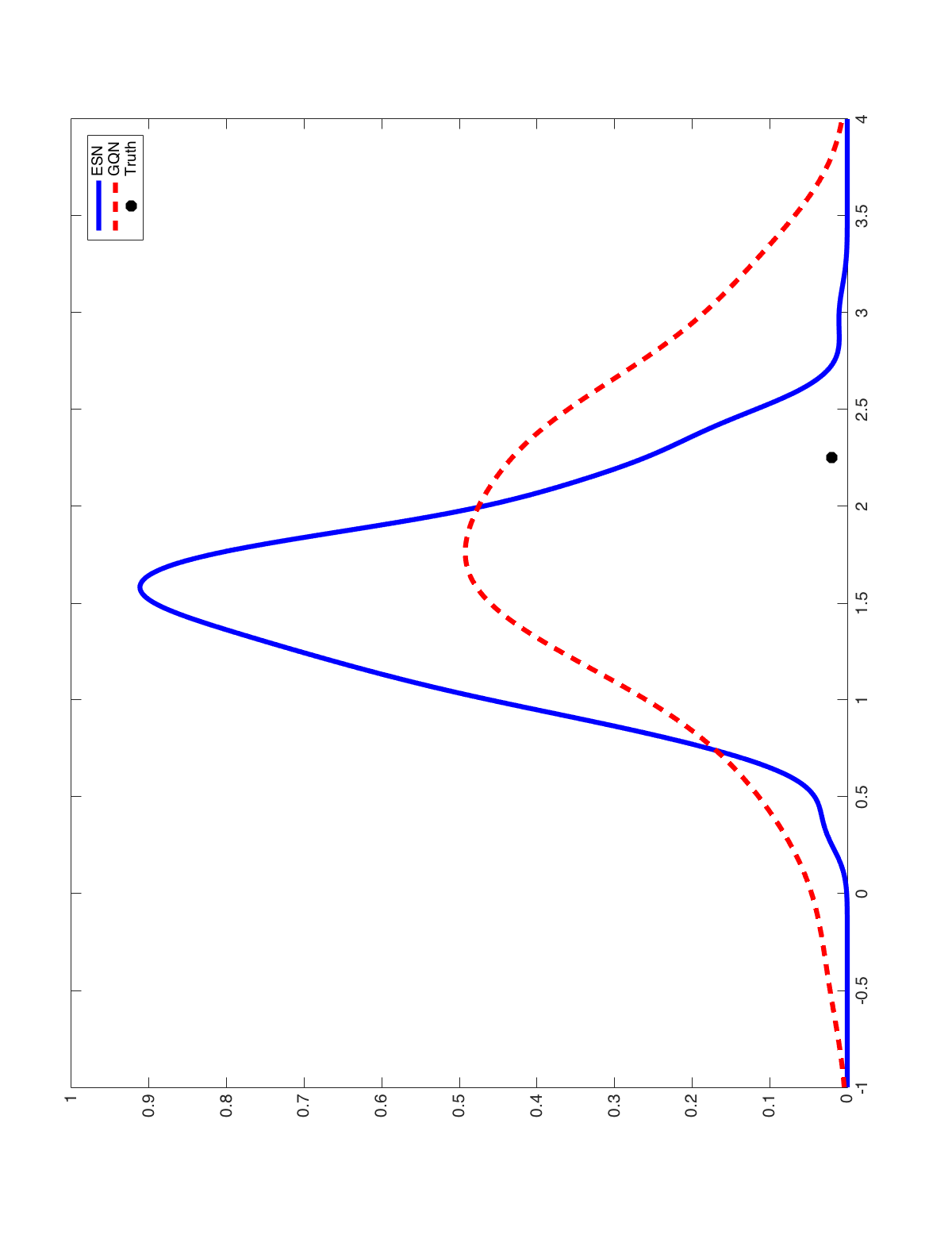} &
   \includegraphics[width=4.5cm,angle=-90]{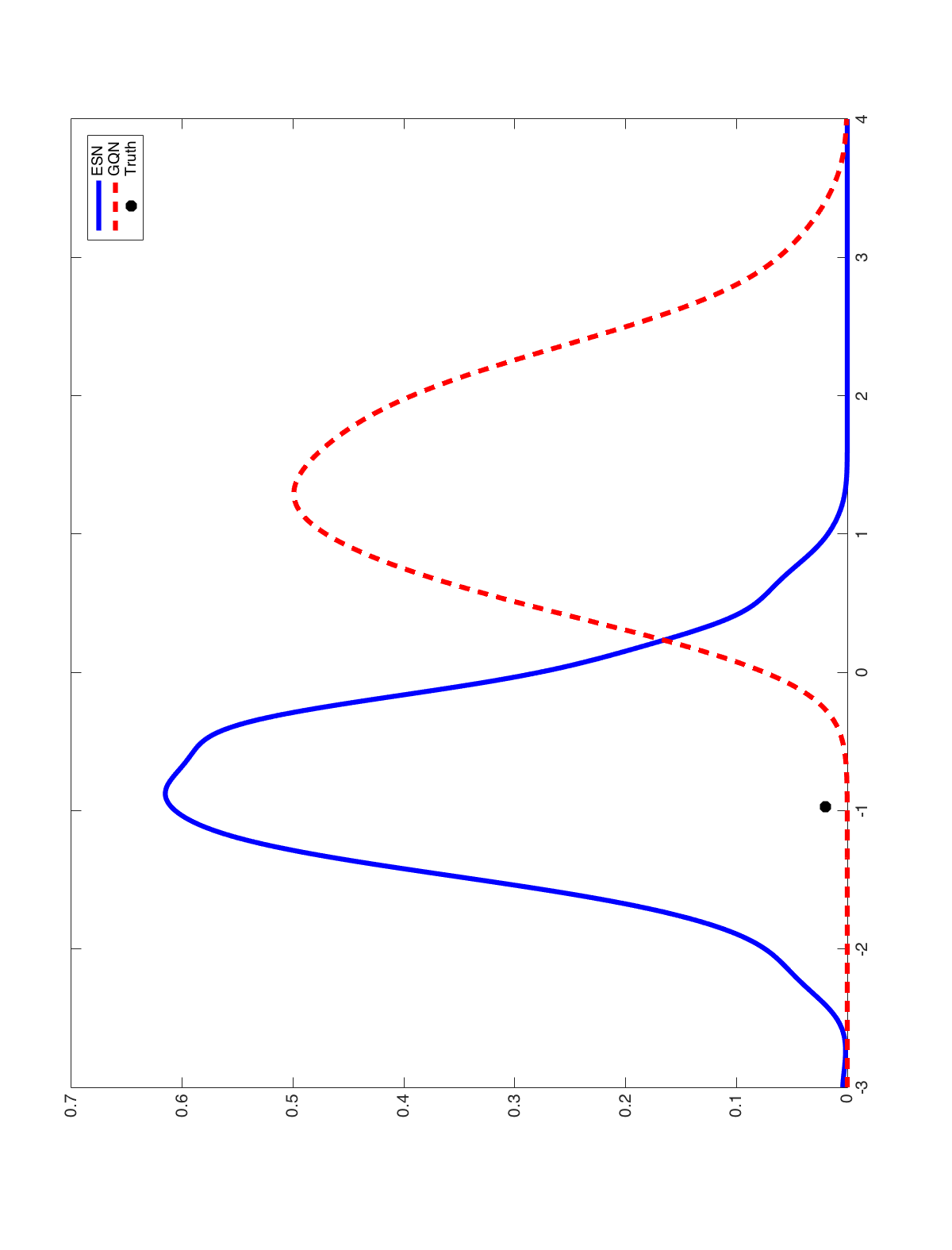}
   \end{tabular}
\caption{Comparison of ensemble QESN 6-month lead forecasts of SST averaged over the the Ni\~no 3.4 Index region to the General Quadratic Nonlinearity (GQN; \cite{wikle2010general}) model forecasts. Left panel: Forecast for October 2015; Right panel: Forecast for October 2016.  The solid blue line represents the kernel smooth of samples from the ESN model and the dashed red line represents the kernel smooth of samples from the GQN model.  The dark circle corresponds to the value of the truth index for the verification time.  }  
\label{fig:ESN_GQN}
\end{figure}

The distributional comparison for the Ni\~no 3.4 region averages is shown in Figure \ref{fig:ESN_GQN}.  The left panel shows the forecast distribution for October 2015 (El Ni\~no) and the right panel shows the forecast distribution for October 2016 (La Ni\~na).  In the case of the El Ni\~no period forecast, both model forecast distributions contain the true value, but the GQN model forecast central tendency is closer to the truth than the ensemble QESN, suggesting it was a better forecast distribution.  However, the La Ni\~na forecasts tell a much different story.  The GQN forecast distribution shows a substantial warm bias (and does not include the truth), whereas the ensemble QESN forecast distribution is quite good, with its central tendency close to the truth and a reasonable uncertainty range.  Previously published examples of GQN long-lead forecasts for earlier ENSO events have shown that it tends to perform better for the El Ni\~no phase than it does for the La Ni\~na phase, most likely due to the fact that the El Ni\~no phase evolution is more nonlinear (although, in the past, it has performed better than it did here for the La Ni\~na case; \cite{wikle2010general}, \cite{wikle2011polynomial}, \cite{gladish2014physically}).  It is very encouraging that the ensemble ESN model shows high quality forecast distributions for both periods for the 2015--2016 ENSO, especially given how efficiently it can be computed.

\section{Discussion and Conclusion}

Many spatio-temporal processes are complex dynamical systems with nonlinear interactions across process components and spatio-temporal scales of variability.  Depending on the scales of interest, it is often crucial to include these nonlinear interactions in forecasting situations in order to get realistic predictions and uncertainty measures.  Only relatively recently have parametric statistical models been developed for nonlinear spatio-temporal processes.  These models tend to have quadratic nonlinear structure, ``deep'' (multi-level) parameter levels, and have been shown to be flexible and useful in a variety of applications with large datasets.  However, in these situations, the models can be quite expensive to implement in terms of computational resources.  As an alternative, we consider a spatio-temporal extension of the echo state networks that have been used in the machine learning context for nonlinear time series applications, in order to exploit their efficient reservoir computing framework.  

In particular, for our spatio-temporal extension, we consider three simple modifications of the traditional ESN, which we call the ensemble quadratic ESN, or ensemble QESN.  This includes the addition of embeddings in the input vector, a quadratic term in mapping the hidden state to the response vector, and an ensemble implementation that accounts for the uncertainty associated with the fixed, yet randomly generated, reservoir weight matrices.  We show that these components allow for reasonable forecasts of nonlinear spatio-temporal processes at a fraction of the computational time associated with more formal statistical methods that accommodate realistic uncertainty quantification.  This is demonstrated on both a classic simulated nonlinear system from \cite{Lore:96} and the challenging problem of long-lead forecasting of Pacific SST during the most recent intense ENSO cycle. 

Although the results presented here are encouraging in the quality of the forecasts relative to the computation time, there remain issues related to the choice of model hyper-parameters.  That is, we rely on a validation sample, drawn from a contiguous time period in the past (e.g., the 1997-1998 ENSO events) to obtain these hyper-parameters.   If these methods are to be used for spatio-temporal forecasting in practice, then it would be useful to consider more formally the uncertainty associated with these choices as well.  In addition, it is of interest to consider this simple methodology on a wider range of spatio-temporal prediction problems and to consider its utility in spatio-temporal classification problems.

In conclusion, with slight modification, relatively simple ``off-the-shelf'' machine learning methods for complex sequential data can be effective in spatio-temporal prediction.  This suggests that other ``black box'' learning tools for dependent data may also be useful to help motivate more formal statistical models for spatio-temporal data.

\section*{Acknowledgements}

This work was partially supported by the U.S. National Science Foundation (NSF) and the U.S. Census Bureau under NSF grant SES-1132031, funded through the NSF-Census Research Network (NCRN) program.

\newpage
\bibliography{reference}
\bibliographystyle{asa}

\end{document}